\documentclass[12pt]{article}

\usepackage{sbc-template}

\usepackage{graphicx,url}

\usepackage[utf8]{inputenc}  

\usepackage{amsfonts}

\usepackage{hyperref}
     
\title{Beyond Random Sampling: Instance Quality-Based \\ Data Partitioning via Item Response Theory}

\author{Lucas Cardoso\inst{1,2}, Vitor Santos\inst{1}, Ricardo Prudêncio\inst{4}, José Ribeiro\inst{3}, \\ Regiane Kawasaki\inst{1} and Ronnie Alves\inst{1,2}}

\address{Universidade Federal do Pará (UFPA)\\ Belém -- PA -- Brasil
\nextinstitute
    Instituto Tecnológico Vale (ITV)\\ Belém -- PA -- Brasil
\nextinstitute
    Instituto Federal do Pará (IFPA)\\ Belém -- PA -- Brasil
\nextinstitute
    Universidade Federal de Pernambuco (UFPE)\\ Recife -- PE -- Brasil
    \email{lucas.cardoso@icen.ufpa.br, vitor.cirilo3@gmail.com, rbcp@cin.ufpe.br}
    \email{jose.ribeiro@ifpa.edu.br, kawasaki@ufpa.br, ronnie.alves@itv.org}
}

\begin{document} 

\maketitle

\begin{abstract}
Robust validation of Machine Learning (ML) models is essential, but traditional data partitioning approaches often ignore the intrinsic quality of each instance. This study proposes the use of Item Response Theory (IRT) parameters to characterize and guide the partitioning of datasets in the model validation stage. The impact of IRT-informed partitioning strategies on the performance of several ML models in four tabular datasets was evaluated. The results obtained demonstrate that IRT reveals an inherent heterogeneity of the instances and highlights the existence of informative subgroups of instances within the same dataset. Based on IRT, balanced partitions were created that consistently help to better understand the tradeoff between bias and variance of the models. In addition, the guessing parameter proved to be a determining factor: training with high-guessing instances can significantly impair model performance and resulted in cases with accuracy below 50\%, while other partitions reached more than 70\% in the same dataset.
\end{abstract}
     

\section{Introduction}

The growing application of Machine Learning (ML) algorithms in diverse areas, from finance and marketing to engineering and biotechnology, drives significant advances in the automation of complex tasks and in the extraction of knowledge from large volumes of data \cite{jordan2015machine}. However, the performance and reliability of these models are intrinsically dependent on the quality and characteristics of the input data \cite{fan2013mining}. Challenges such as instance heterogeneity, class imbalance \cite{he2009learning} and the presence of noise or unpredictability \cite{aggarwal2017introduction} are common in many real-world datasets. Such conditions can compromise the generalization capacity of models and the confidence in their predictions.

Model validation is a critical step during ML development. Although techniques such as stratified cross-validation are widely used, they are typically based on random partitions. Such approaches, however, do not consider the intrinsic quality of each individual instance. This gap in assessing the inherent properties of instances in the data partition was what motivated the present study.

To address this limitation, this study proposes a validation methodology that uses Item Response Theory (IRT) \cite{baker2001basics} as a tool to characterize the quality of instances. For this, the parameters that describe the IRT item were used: Discrimination, Difficulty, and Guessing. Based on these characterizations, training and test set partitioning strategies that go beyond randomness were developed and evaluated in order to create more controlled learning and evaluation scenarios to test ML models.

Although IRT originated from psychometrics and is traditionally used to assess individuals, such as in the ENEM (National High School Exam) \cite{enem}, applied to students who are finishing high school, several recent studies \cite{prudencio2015analysis,martinez2019item} have already demonstrated that a simple analogy is sufficient to use the assessment potential of IRT applied to ML. By considering the models as individuals and the instances of the dataset as test items, it is possible to use the robust assessment tools that IRT has. Such studies have already demonstrated that IRT has an informative potential that is still little explored in the field of machine learning.

This work investigates the impact of these IRT-guided partitioning strategies on the performance of several ML models generated from different families of learning algorithms. As a case study, four public and well-established tabular datasets in the literature (\textit{Heart-statlog}, \textit{Ilpd}, \textit{Diabetes} and \textit{Breast-w}) were chosen, which present common challenges of heterogeneity and imbalance. The results obtained demonstrated that the application of IRT principles allows a deeper understanding of data heterogeneity. Furthermore, it was observed that ``balanced'' partition strategies, which distribute the complexity of the instances in a balanced way, on several occasions generated results that surpassed random approaches. In addition, it was identified that the guessing parameter, which indicates the level of unpredictability of the instances, can be a significant factor in the drop in model performance when not managed properly. Although it is not the focus of the work, there is a proximity to the bias-variance dilemma, since the specific combinations of instances for training and testing had an impact on the behavior of the models and, at times, increased or decreased the bias-variance.

The remainder of this paper is organized as follows: Section \ref{sec:background} presents the theoretical framework on the use of Item Response Theory in the proposed context. Section \ref{sec:methodology} details the methodology used to calculate the item parameters and to create the partition strategies informed by IRT. Section \ref{results} presents and discusses the results obtained in each dataset analyzed. Finally, Section \ref{conclusao} concludes the work and suggests future research.

\section{Background} \label{sec:background}


This section will contextualize how IRT works and how it can be applied in a machine learning context. Originating in psychometrics, Item Response Theory was developed as a more robust way to evaluate individuals responding to a test. The difference between IRT and ML is its \textit{instance-centric} nature. Unlike the classical batch-based evaluation method, which is commonly used, IRT focuses on the items and not the entire test. It consists of several mathematical models that aim to evaluate a respondent's performance individually for each item in the test. It commonly considers the dichotomous nature of the item, i.e., it only evaluates whether the item was answered correctly or not \cite{baker2001basics}. Within ML, this occurs when considering whether a given instance was classified correctly or not.

In order to evaluate the items individually, IRT uses the parameters that describe the item, the three main ones being: Discrimination, Difficulty and Guessing. Each parameter is represented by a numerical value and represents a specific characteristic of the item \cite{de2000teoria}. Discrimination $a_{i}$ characterizes the item in terms of its ability to discriminate between high-ability and low-ability respondents, i.e., it represents how well the item serves to discriminate the individuals evaluated based on the result of their response. So, if the individual answers the item correctly then he probably has high ability, if not then he probably is a low-ability respondent \cite{de2000teoria}. In ML, this would be an instance that has the ability to identify which model is more suitable based on its accuracy, e.g., for a scenario with two models with statistically very similar performance, the number of correct answers in very discriminative instances can be used to choose the best model.

The Difficulty $b_{i}$ is the direct measurement of how difficult a given item is to answer correctly. The higher its value, the greater the skill required to answer it correctly \cite{de2000teoria}. In ML, a high difficulty instance may represent a condition that was little explored during the training stage and that the model was unable to generalize well enough. Theoretically, the range of the difficulty and discrimination parameters is the set of real numbers $\mathbb{R}$ $(-\infty,+\infty)$.

The Guessing parameter $c_{i}$ is a little different, it is not represented by a real number $\mathbb{R}$, but rather by a probability that can vary from $0.0$ to $1.0$. Guessing is the chance of a random hit, i.e., it is the chance of a certain item being answered correctly by chance, it also happens when a low-skill respondent gets a difficult item right \cite{baker2001basics}. In ML, the high guessing value can be a representation of randomness, where a model tends to choose any class for these instances.

The characterization of instances by item parameters can reveal information that is still little explored on a large scale in the ML universe. Several recent studies have already explored IRT together with ML to evaluate specific challenges that exist in ML, for example: \cite{cardoso2020decoding, song2021efficient} used IRT to evaluate benchmarks; \cite{araujo2023quest} used IRT to evaluate and determine which models are suitable for production; \cite{de2024explanations} created a new methodology based on IRT to generate explanations of model prediction; \cite{cardoso2024standing} combined IRT with the confusion matrix to create a more refined form of evaluation. The present study, in turn, proposes to use IRT as a tool to explore the selection of instances for training and evaluation of models.

\section{Methodology} \label{sec:methodology}

\begin{figure}[!ht]
\centering
\includegraphics[width=0.9\textwidth]{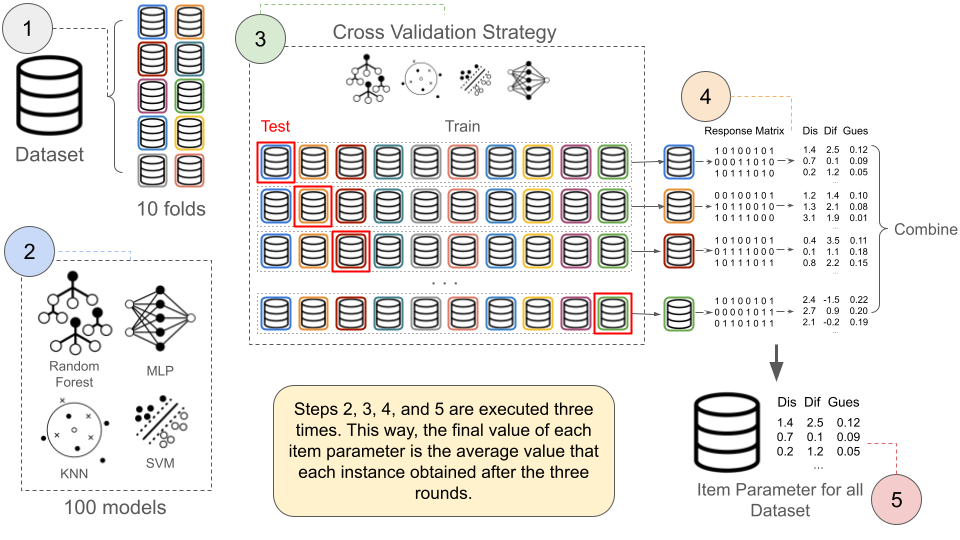}
\caption{Flowchart for calculating item parameters.}
\label{fig:meth_1}
\end{figure}

\begin{figure}[!ht]
\centering
\includegraphics[width=0.9\textwidth]{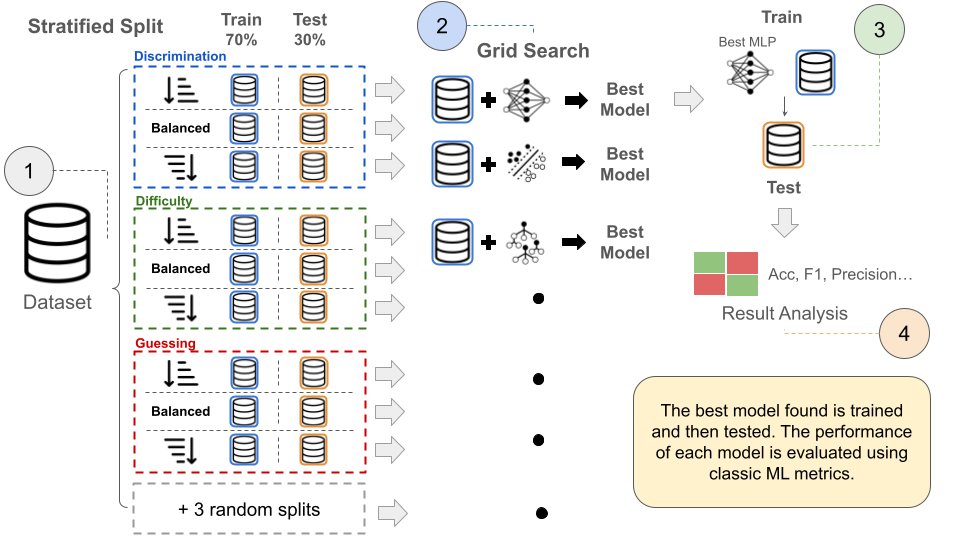}
\caption{Flowchart for creating data partitions and evaluating models.}
\label{fig:meth_2}
\end{figure}

This section aims to explain the methodology used to achieve the objective of evaluating the impact of different training and testing sets defined from a non-arbitrary choice, i.e., from the IRT item parameters. The methodology \footnote{All source code for the methodology can be accessed at: \href{https://github.com/LucasFerraroCardoso/IRT_data_partition}{\texttt{https://github.com/LucasFerraroCardoso/IRT\_data\_partition}}} was divided into two main parts: (1) Calculation of the item parameters for all instances of the chosen dataset and (2) Creation of different data partitions based on the value of the item parameters and evaluation of the models. Figure \ref{fig:meth_1} presents the flowchart of Part 1 of the methodology, while Figure \ref{fig:meth_2} presents Part 2.

\noindent (1) Calculation of item parameters for all instances of the dataset.

\begin{enumerate}
    \item The dataset is divided into 10 stratified folds;
    \item 100 random models from 10 different families of algorithms are generated;
    \item Using the cross-validation strategy, each random model will be trained and tested with the 10 folds, thus having 100 responses for each of the 10 folds.
    \item The responses of the models are used to calculate the item parameters for each instance of each fold.
    \item The folds are combined into a single set of instances, thus having the item parameters calculated for all instances of the dataset.
\end{enumerate}

The generation of models from different families was done using the Scikit-Learn library \cite{pedregosa2011scikit}, in total there are models from 10 different algorithms. Tree-based: Decision Tree (DT) and Random Forest (RF); Ensemble: AdaBoost (ADA), Gradient Boosting (GB) and Bagging (BAG); Neural Networks: Multilayer Perceptron (MLP); Based on the distance between neighbors: k-Nearest Neighbors (KNN); Support Vectors: Support Vector Machine (SVM) and Linear Support Vector Machine (LSVM); Linear Discrimination: Linear Discriminant Analysis (LDA). For each algorithm, models with different combinations of hyperparameters (random models) are generated. This is done to generate a wide diversity of responses for each instance of the dataset.

To generate the item parameters, the Ltm library \cite{rizopoulos2006ltm} available for R was used. To do this, it is enough to have a response matrix of the items of interest. The response matrix is a binary matrix where the rows are the models and the columns are the items. The value of each element of the matrix is the dichotomous result of the classification of the models for each item, being 1 in case of a correct response and 0 otherwise.

\noindent (2) Creation of different data partitions based on the value of the item parameters and evaluation of the models.

\begin{enumerate}
    \item The dataset is divided based on the item parameters, for this the instances are ordered in ascending and descending order for each parameter. After each ordering, the first 70\% instances will be part of the training set while the remaining 30\% will form the test set. The ordering is always done separately for each class of the dataset, thus ensuring correct stratification. A balanced division is also created, where the 70/30 partitions are composed of instances with low, medium and high values of the item parameter. In addition, 3 random splits are also created for comparison.
    \item For each data partition, either by item parameter or random, a Random GridSearch is performed with only the training data to find the best model from each of the algorithm families. This way, there will be a best model of each family for each of the partitions created.
    \item The best model found for each family is trained and then tested with the training and testing data of its partition.
    \item Finally, the performance of each model is then analyzed with the classic ML metrics: Accuracy, F1 score, Precision, Recall and MCC.
\end{enumerate}

By creating multiple partitions based on item parameters, the goal is to create theoretically extreme conditions to compare the behavior of models when trained and tested under these conditions.

\subsection{Datasets} \label{datasets}

To carry out the study proposed in this research, 4 well-known binary datasets were selected as use cases. Table \ref{tab:tab1} lists the main characteristics of each dataset. All the chosen datasets are from a medical context: \textit{ilpd} is a liver disease dataset, \textit{heart-statlog} is a heart disease dataset, \textit{diabetes} is self-explanatory and \textit{breast-w} is a breast cancer dataset. All datasets can be obtained for free on the OpenML platform \cite{vanschoren2014openml}.

Datasets from a medical context are naturally complex. Since they are patient diagnoses, they commonly have heterogeneity of instances, imbalance and presence of noise. Furthermore, because it is a sensitive context, the accuracy and reliability of the ML models generated from them are of utmost importance for clinical decision-making by a physician. These characteristics make these datasets interesting pieces for evaluating the proposed methodology.

\begin{table}[!ht]
\centering
\caption{List of datasets.}
\label{tab:tab1}
\begin{tabular}{|l|l|l|l|l|l|}
\hline
Datasets      & Nº Classes & Nº Instances & Nº Features & \% Majority & \% Minority \\ \hline
ilpd          & 2          & 583          & 10          & 71.35       & 28.65       \\ \hline
heart-statlog & 2          & 270          & 13          & 55.56       & 44.44       \\ \hline
diabetes      & 2          & 768          & 8           & 65.10       & 34.90       \\ \hline
breast-w      & 2          & 699          & 9           & 65.52       & 34.48       \\ \hline
\end{tabular}
\end{table}

\section{Results and Discussion} \label{results}

This study aims to investigate the impact of different training and testing sets, defined from IRT estimators, on the performance of different ML models. As a case study, four medical datasets were used: \textit{ilpd}, \textit{heart-statlog}, \textit{diabetes} and \textit{breast-w}. The results obtained demonstrate that the intrinsic properties of the instances of a dataset, revealed by the item parameters, can significantly influence the performance of the models, where certain partition strategies can outperform classical random approaches, while others can drastically impair the learning of the models. In this section, the results obtained by the experiments performed and the discussions held about them will be presented \footnote{All executed source code and generated files can be found in the repository available at \href{https://github.com/LucasFerraroCardoso/IRT_data_partition}{\texttt{https://github.com/LucasFerraroCardoso/IRT\_data\_partition}}}.

\subsection{Characterization of Datasets via IRT Item Parameters}

The analysis of the arrangement of item parameters for each dataset, obtained after executing step 1 of the methodology (Section \ref{sec:methodology}), alone reveals an important piece of information: all datasets presented intrinsic heterogeneity of their instances. This means that within a dataset there are natural differences in the quality of the information that each sample represents during a machine learning task. That is, not all instances of a dataset can be considered equally complex, difficult or informative for a model, but rather that there are several subgroups of instances that models can treat in different ways, even though they belong to the same class.


\begin{figure}[ht]
    \centering
    \begin{minipage}{0.45\textwidth}
        \centering
        \includegraphics[width=\textwidth]{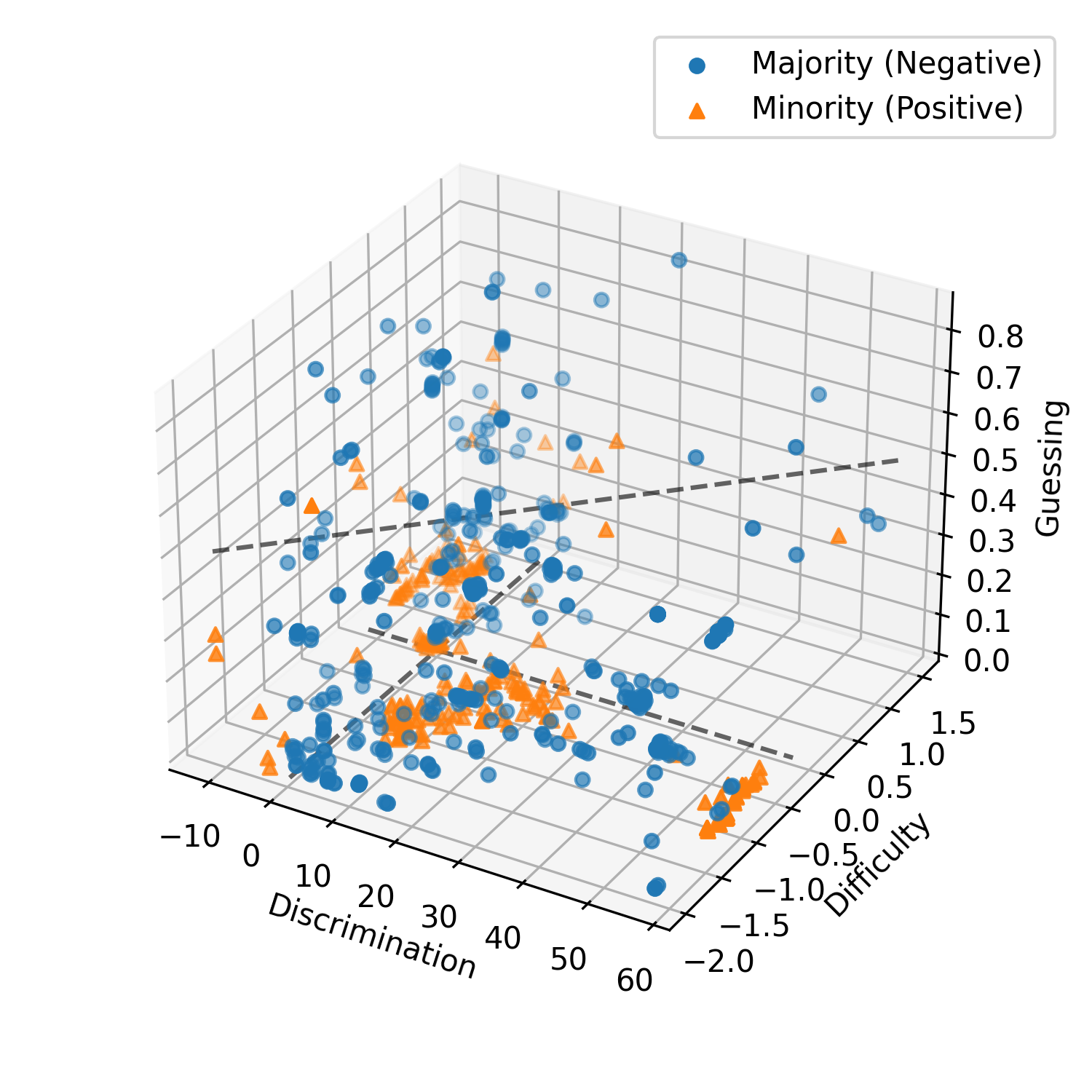}
        (a) Dataset breast-w.
    \end{minipage}\hfill
    \begin{minipage}{0.45\textwidth}
        \centering
        \includegraphics[width=\textwidth]{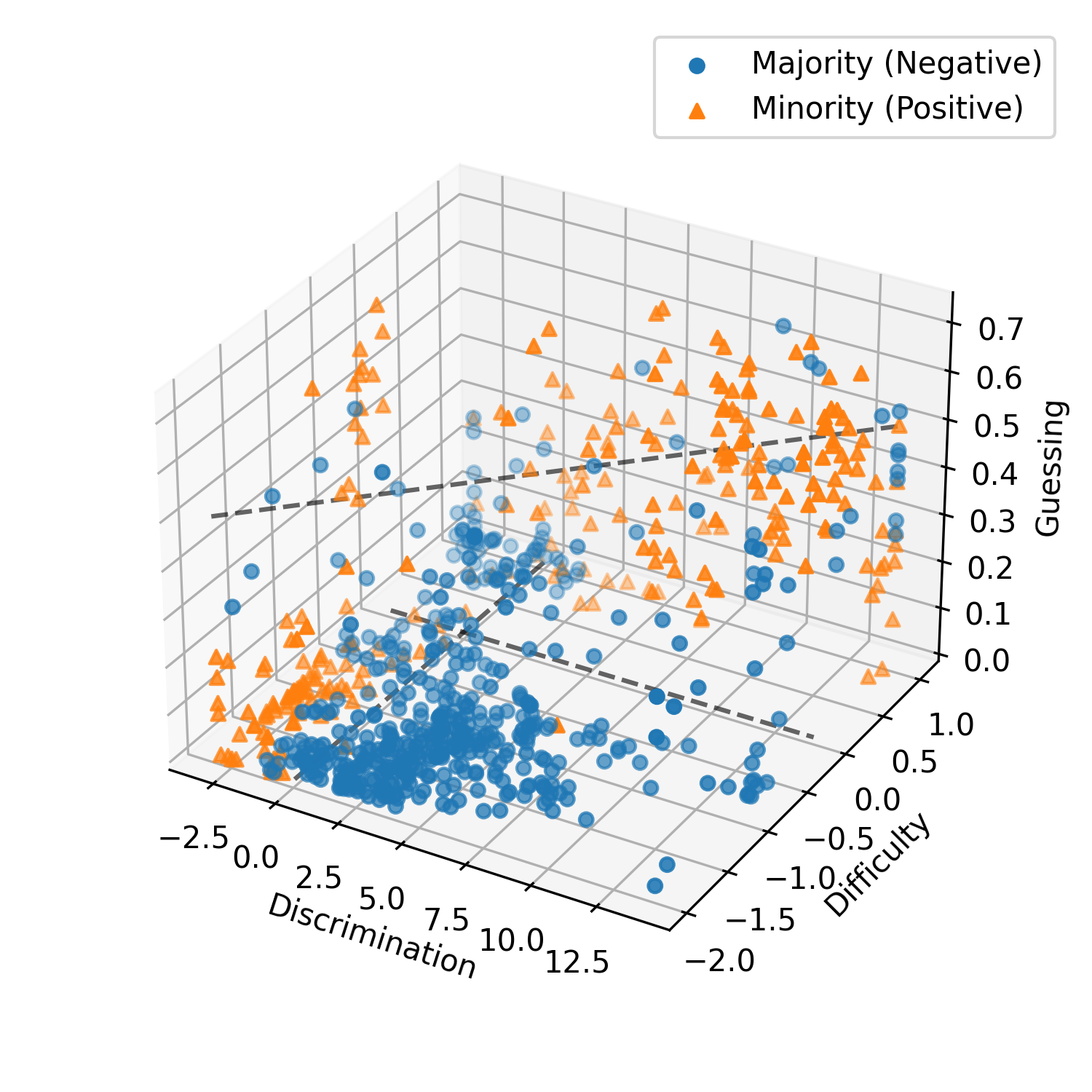}
        (b) Dataset diabetes.
    \end{minipage}
    \caption{Three-dimensional arrangement of item parameters.}
    \label{fig:plot3d}
\end{figure}

It can be observed in Figure \ref{fig:plot3d} that even instances of the same class can be grouped in different ways by item parameters. For example, for the dataset \textit{diabetes}, which has average discrimination values equal to $3.82$, difficulty equal to $-0.69$ and guessing equal to $0.15$, it is clear that there are two subgroups of instances of the minority class, one subgroup that has low values of discrimination, difficulty and guessing, while the other has high values for all parameters. For the dataset \textit{breast-w}, which has an average discrimination of $12.75$, average difficulty of $-0.35$ and average guessing of $0.19$, it is also possible to note the existence of subgroups, e.g. for the minority class it is clear that some instances have high difficulty and low discrimination value, while other instances present the opposite condition, low difficulty value and high discrimination.

Thus, depending on the subgroup of instances that persist in the training or testing set, it is expected that this will influence the final result of the model, increasing the variance. Furthermore, is there a better subgroup? Or one that better represents the data that the model is expected to be able to classify? Some instances may be very close to the decision limit or present ambiguous characteristics that hinder the model's learning. Studies such as \cite{cardoso2024standing} have already explored how negative discrimination can be a thermometer to find these instances. To answer these questions, a possible future work would be to explore more specifically the impact of these instances on learning and their informative relevance for the dataset. In this work, the impact of combinations of instances with different item parameter values is explored globally.

\subsection{Impact of Partition Strategies on Model Performance}

\begin{figure}[!ht]
\centering
\includegraphics[width=0.7\textwidth]{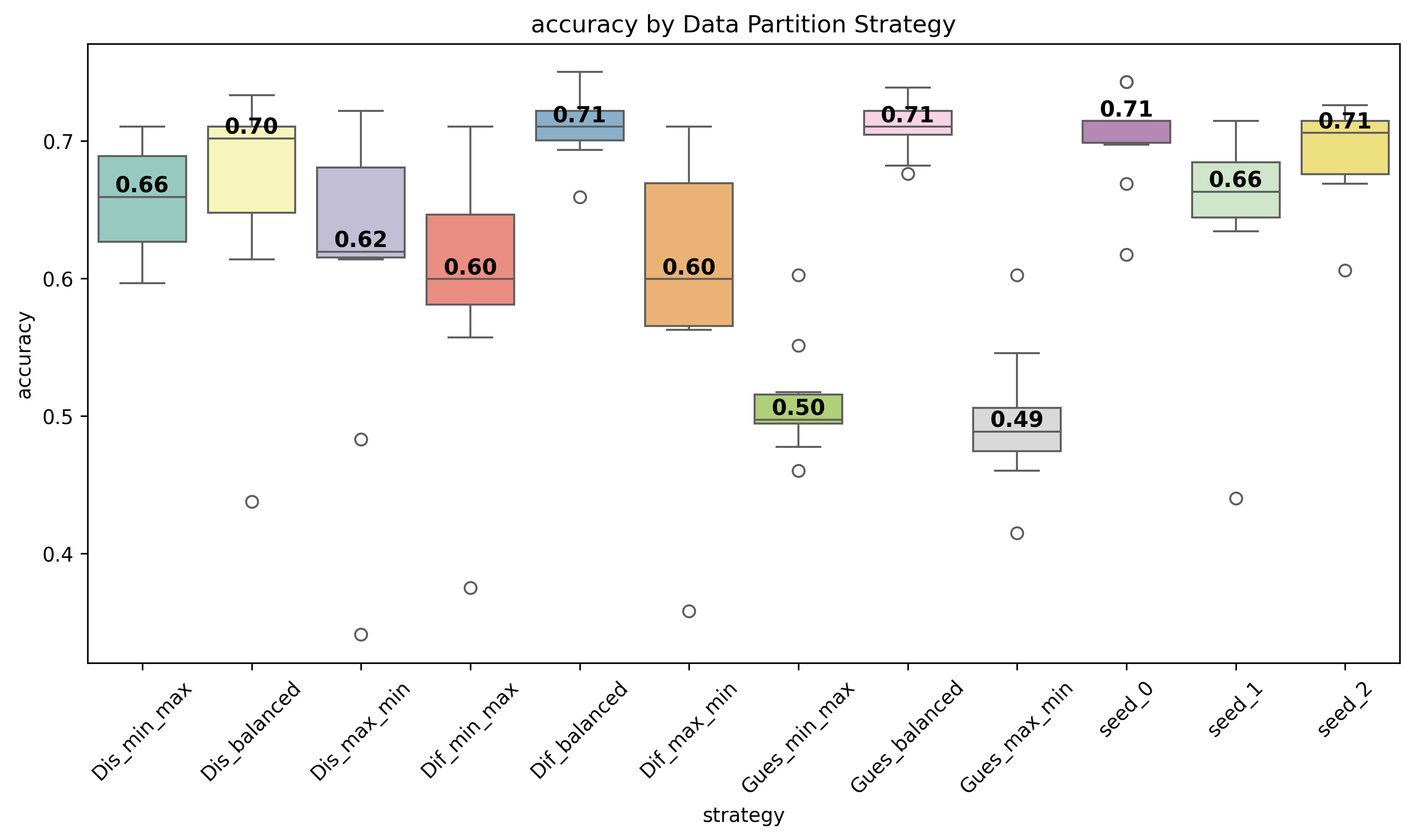}
\caption{Accuracy Distribuition of ilpd.}
\label{fig:box-plot_ilpd}
\end{figure}

The evaluation of the overall performance of the partition strategies confirmed the existence of an impact: data partitions chosen in non-arbitrary ways can improve or harm the final performance of the models. This is easily observed in Figure \ref{fig:box-plot_ilpd}, where for the \textit{ilpd} dataset the best average accuracy scores were obtained by partitions chosen based on the item parameters, in which the \textit{Dif\_balanced} and \textit{Gues\_balanced} strategies stand out not only for the highest average score (approximately $0.71$), but also for presenting flattening of their respective box-plots, lower variance between the models.


It is also noted that the \textit{Dif\_balanced} and \textit{Gues\_balanced} strategies presented interesting results that are comparable to purely random executions (seeds 0, 1 and 2). Given this, it can be assumed that this approach, which ensures the proportional representation of instances of all ranges of item parameters in both training and test sets (along with class stratification), provides a more robust learning and evaluation environment. By exposing the model to the full range of complexity of the dataset, it is able to learn more generalizable patterns. Furthermore, the flatness and high mean value of the box-plot reveal a respective decrease in the variance and bias of the models. Another interesting result is the very low scores obtained with the \textit{Gues\_min\_max} and \textit{Gues\_max\_min} partitions ($0.51$ and $0.50$, respectively), which aim to create extreme cases where the models are deliberately trained with the lowest guessing instances and tested with the highest guessing instances and vice versa.

The \textit{Gues\_max\_min} strategy stands out even more, as the models often obtained accuracy lower than 50\%. High guessing values can be indicators of randomness, meaning that the models are unable to determine a pattern to correctly classify this group of instances and in the end the model just chooses a class arbitrarily. According to IRT concepts, items with a high randomness rate are complex to be clearly evaluated, for ML a noise or outlier can be the cause of high guessing. The results of these partitions indicate that introducing substantial noise or inherently unpredictable instances into the training set fundamentally compromises the model's ability to learn meaningful patterns, resulting in models that learn the noise and fail to generalize. Both extreme guessing-based strategies were able to decrease variance (flattening of the box plot), but they increased bias (low mean).

\begin{figure}[!ht]
\centering
\includegraphics[width=0.8\textwidth]{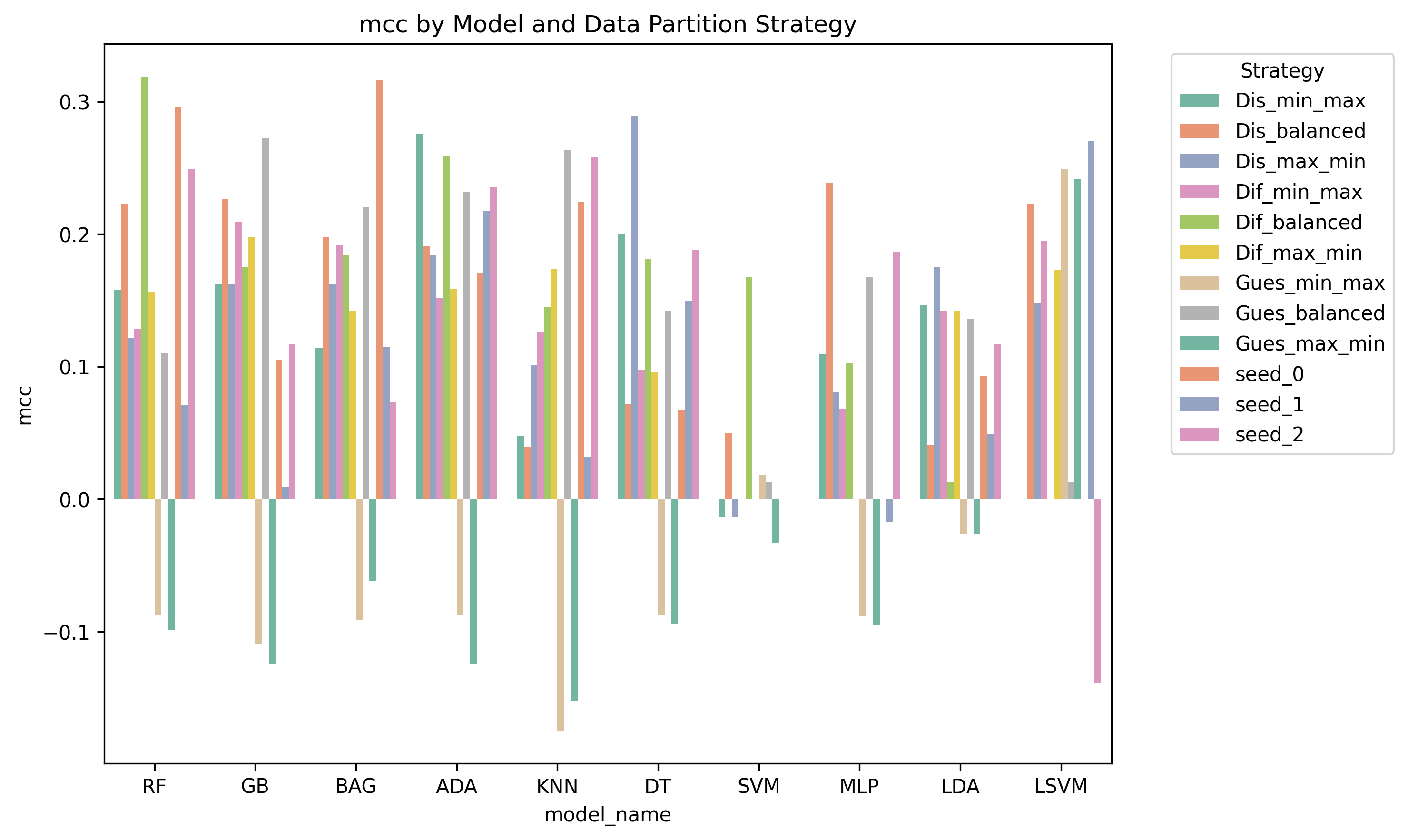}
\caption{MCC score obtained by models in ilpd.}
\label{fig:bar-plot_mcc_ilpd}
\end{figure}

These observations are reinforced by the individual results of each model with the MCC metric, which is traditionally used for evaluation in the context of class imbalance, as is the case with \textit{ilpd}. Figure \ref{fig:bar-plot_mcc_ilpd} shows the MCC score for all models in all partitions created for the dataset. It can be seen that the highest scores obtained by the models are mostly in the \textit{IRT\_balanced} partitions, while the extreme partitions based on guessing result in negative values. The MCC is a metric that varies from $-1$ (very bad) to $1$ (excellent), with negative values indicating that the model is not only making mistakes, but inverting the classifications.

\begin{figure}[!ht]
    \centering
    \begin{minipage}{0.5\textwidth}
        \centering
        \includegraphics[width=\textwidth]{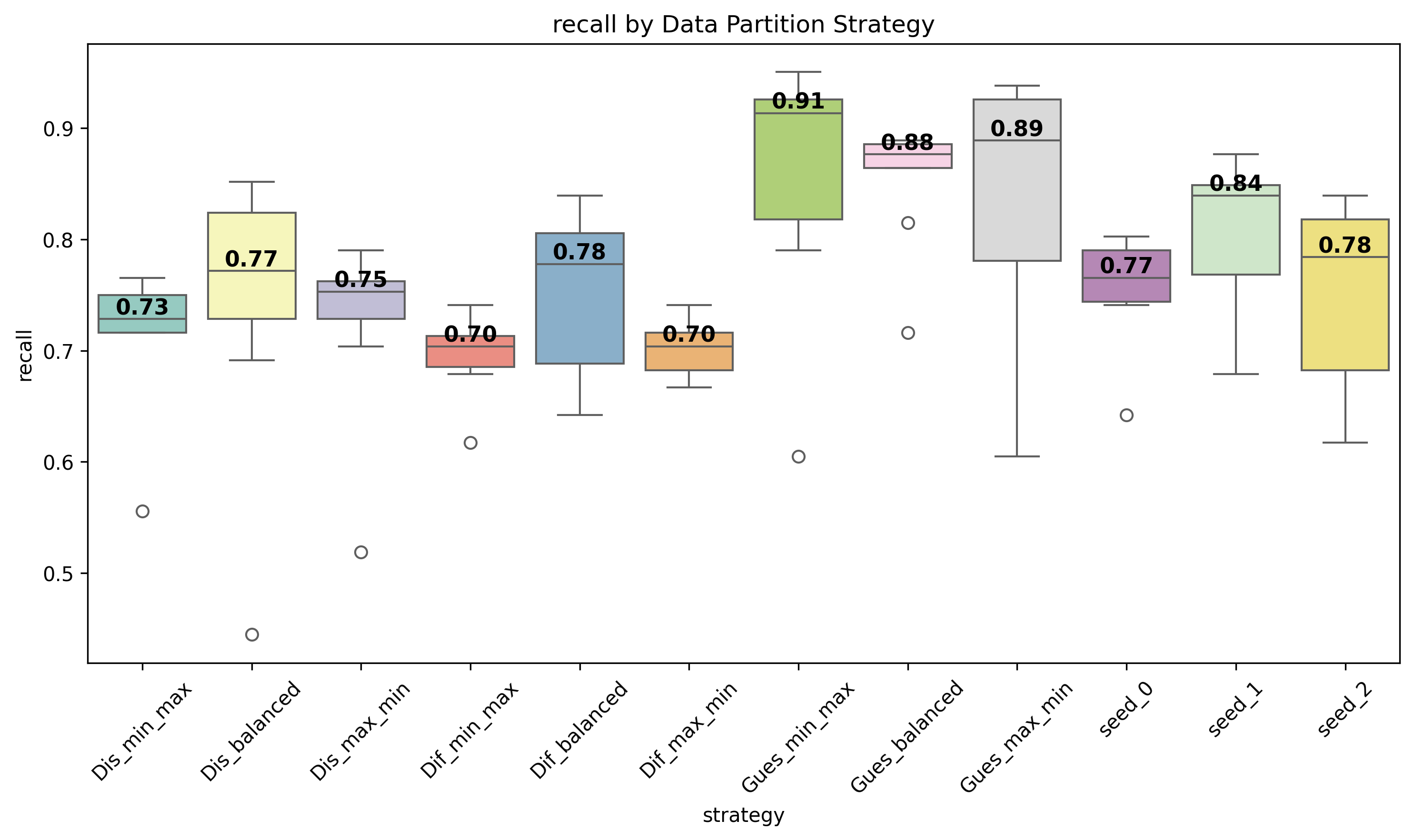}
        (a) Recall variation.
    \end{minipage}\hfill
    \begin{minipage}{0.5\textwidth}
        \centering
        \includegraphics[width=\textwidth]{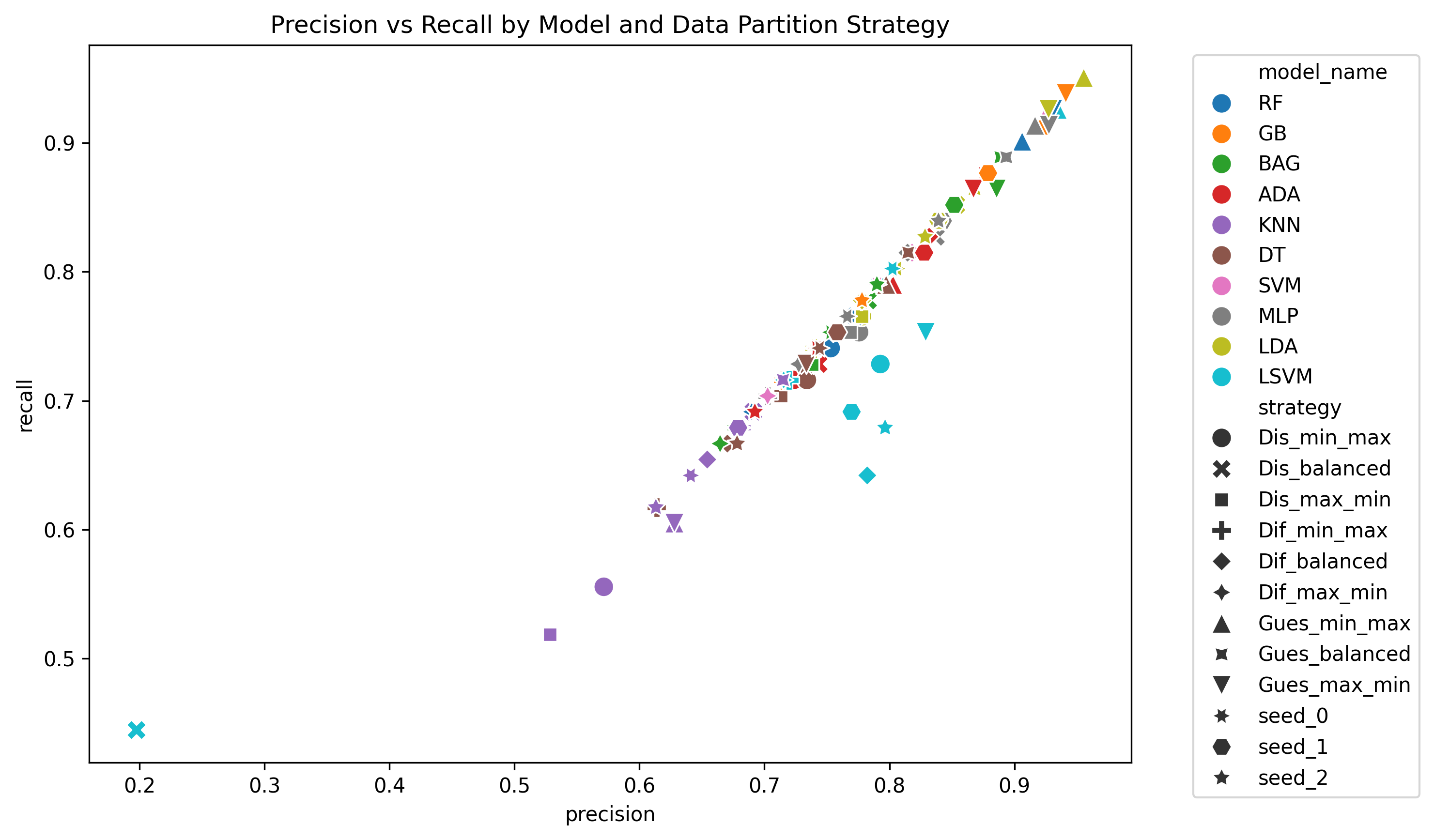}
        (b) Recall x Precision.
    \end{minipage}
    \caption{Recall and Precision score obtained by models in heart-statlog.}
    \label{fig:box-plot_recall_heart}
\end{figure}

Another interesting result was obtained for the \textit{heart-statlog} dataset; Figure \ref{fig:box-plot_recall_heart} (a) shows the variation of the Recall metric obtained for the dataset. In medical contexts, metrics such as Recall and Precision are fundamental to measure the model's ability to correctly classify \textit{True-Positives} cases, i.e., cases in which the patient actually has the disease, as errors in this sense can lead to both practical and ethical consequences.


It can be observed that the best results are obtained for data partitions based on guessing, with emphasis on \textit{Gues\_min\_max}. In this case, training with low guessing and testing with high guessing resulted in the highest average score obtained; this suggests that, if the model learns robust patterns from ``clean'' data, then it is possible to assume that it is able to deal with some level of noise in the test. As can be seen in Figure \ref{fig:box-plot_recall_heart} (b), the model with the highest Recall and Precision scores was LDA, with 0.95 for both metrics, obtained in the \textit{Gues\_min\_max} partition strategy. Although this strategy generated the models with the highest performance levels, it is possible to notice a variance in the results due to a large amplitude of the box-plot; in contrast, the \textit{Gues\_balanced} strategy shows a much smaller variance (flattening of the box-plot).

\begin{figure}[!ht]
\centering
\includegraphics[width=0.7\textwidth]{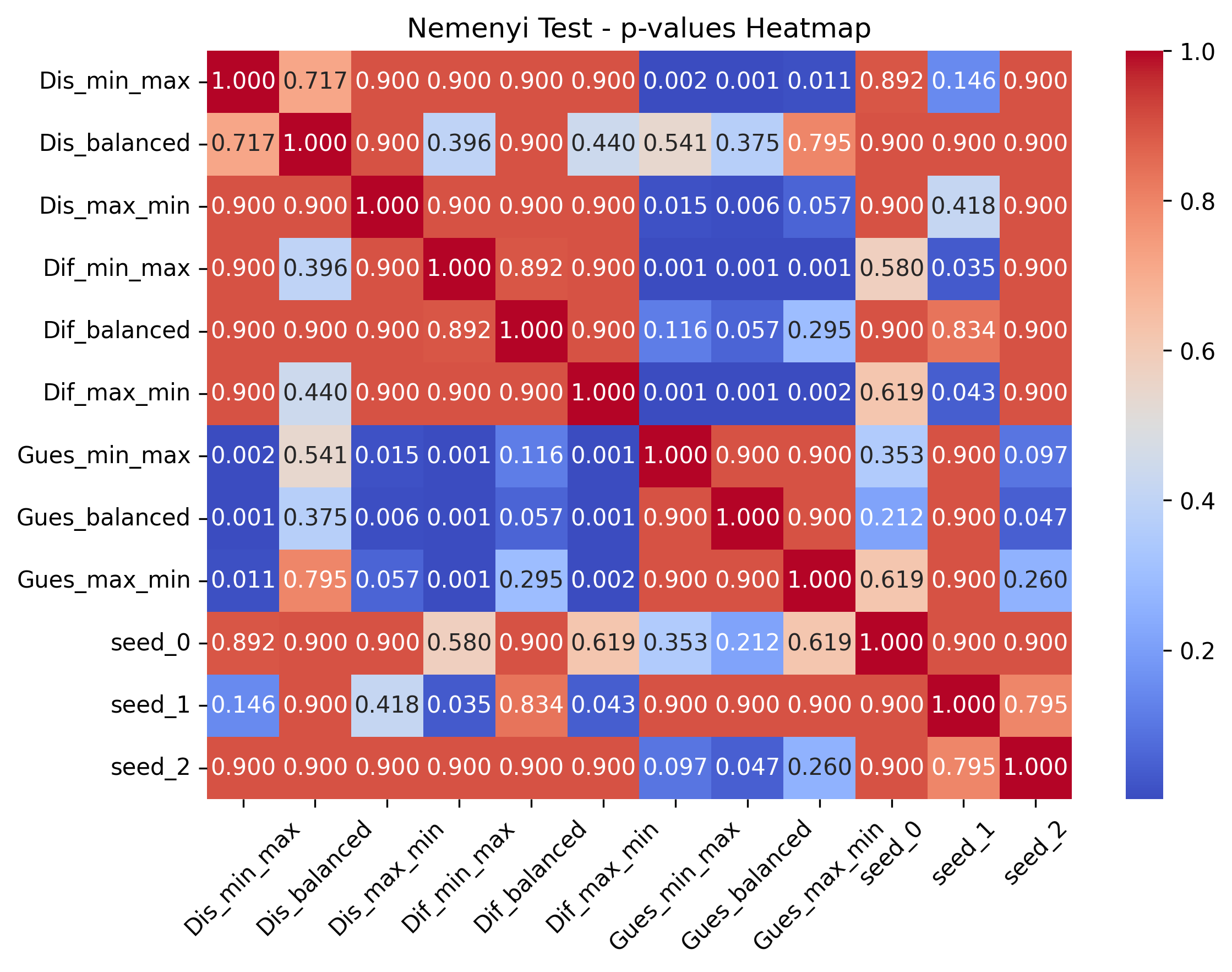}
\caption{Heatmap of Nemenyi test p-values for Heart-Statlog.}
\label{fig:heatmap_nemenyi_heart}
\end{figure}

The observations are corroborated by statistical tests, where the Friedman test performed using the F1 score results of the models indicated statistically significant differences between the partition strategies, with $p-value$ below $0.05$ in all datasets, and \textit{post-hoc} analyses with the Nemenyi test frequently revealed very low $p-values$ in comparisons involving strategies based on the guessing parameter (see Figure \ref{fig:heatmap_nemenyi_heart}).

\begin{figure}[!ht]
\centering
\includegraphics[width=0.8\textwidth]{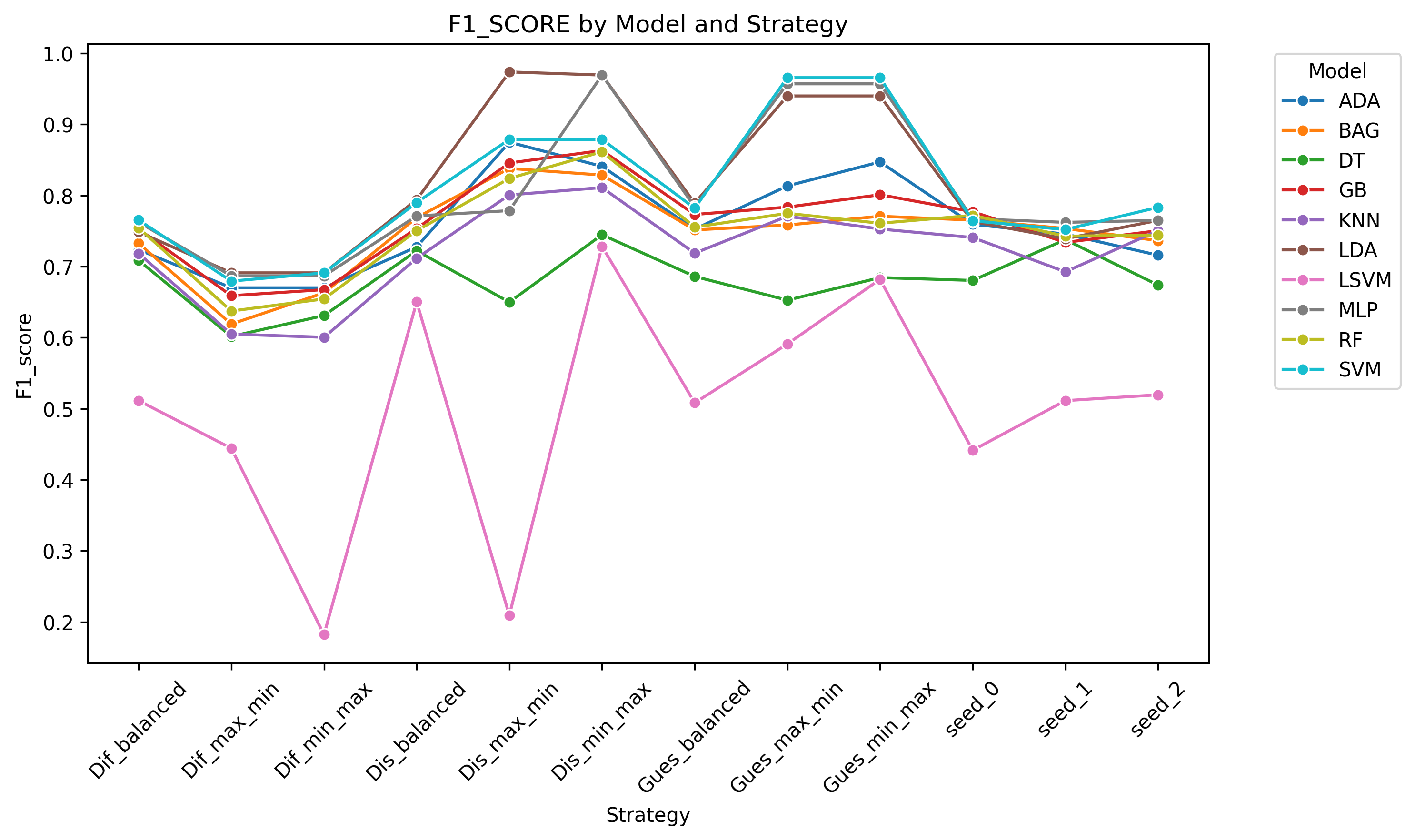}
\caption{F1 score obtained by models in diabetes.}
\label{fig:line-plot_f1_diabetes}
\end{figure}

The study also allows us to analyze the robustness of the evaluated models by analyzing the behavior of the models' performance across different data partitions. Figure \ref{fig:line-plot_f1_diabetes} illustrates this. It shows that models such as SVM and MLP are able to maintain similar F1 values, even across different data partitions. At the same time, models that are known to be weaker, such as LSVM and DT, have more difficulty maintaining their performance across different partitions. LSVM, in particular, stands out negatively with its erratic and very low behavior in some partitions of extreme combinations such as \textit{Dif\_min\_max} and \textit{Dis\_max\_min}. Models with high bias have low generalization capacity. These results confirm that different algorithms may have greater innate generalization capabilities than others. Future work would be to deepen this analysis and explore the behavior of each model individually.

\section{Final Considerations}\label{conclusao}

This study investigates the promising intersection between Item Response Theory (IRT) and Machine Learning, proposing an innovative methodology for creating data partitioning strategies that go beyond randomness. By integrating IRT concepts, it was possible to characterize the instances of a dataset by their psychometric qualities: discrimination, difficulty, and guessing. This approach offers a powerful lens to understand the intrinsic quality of tabular datasets, especially in critical domains such as healthcare, where accuracy and reliability are crucial. The results obtained reveal that the presence of different informative subgroups of instances can directly impact the learning of models. This is where the bias and variance dilemma in machine learning manifests itself most clearly.
Partitions created without considering the quality of the instances can inadvertently introduce or exacerbate this dilemma. For example, we observed that instances with a high guessing value represent a significant challenge for the training environment. Over-inclusion of these instances, which resemble noise or random responses, can lead the model to learn spurious patterns. This translates into either high bias, where the model fails to capture the true relationship between the features and the target, or, on the other hand, high variance, where the model overfits to this noise, losing its ability to generalize to new data. In other words, traditional cross-validation, blind to instance quality, can overestimate or underestimate the model’s performance, leading to erroneous decisions about its robustness.

In contrast, “balanced” partitioning strategies, which ensure subgroups of instances that are balanced in terms of their IRT qualities, reinforce a fundamental lesson: the representativeness of the training and test sets is a more significant factor in the generalization of models than the simple amount of data. By controlling for the psychometric characteristics of the instances in each partition, IRT allows us to mitigate the bias-variance tradeoff more effectively. More balanced partitions tend to reduce the variance in the generalization error estimate, as each cross-validation fold becomes a more faithful representation of the data population. At the same time, by ensuring that the model is not learning from predominantly noisy or low-discrimination data, we can reduce the bias inherent in the learning process, resulting in more robust models that are better able to generalize to unseen data.

As next steps, exploring the use of IRT item parameters in preprocessing steps shows promise. This includes identifying outliers and noise based on their psychometric characteristics, weighting instances during training (giving more weight to high-quality instances and less to instances with low guessing or discrimination value), and developing more sophisticated sampling techniques that focus on the "quality" of instances. Such approaches have the potential to further refine the machine learning process, building models that not only perform well, but also understand and deal with the intrinsic complexity of their data, addressing the bias and variance dilemma in a more granular and strategic way.

\bibliographystyle{sbc}
\bibliography{sbc-template}

\end{document}